\documentclass[letterpaper]{article} 
\usepackage{aaai2026}  
\usepackage{times}  
\usepackage{helvet}  
\usepackage{courier}  
\usepackage[hyphens]{url}  
\usepackage{graphicx} 
\usepackage{natbib}  
\usepackage{caption} 
\usepackage{algorithm}
\usepackage{algorithmic}
\usepackage{amssymb}
\usepackage{amsmath} 
\usepackage{booktabs}
\usepackage{multirow}
\usepackage{xcolor}

\usepackage[table]{xcolor}
\definecolor{myteal}{HTML}{17BEBB}

\usepackage[american]{babel}
\usepackage{microtype}

\usepackage{fontawesome}
\usepackage{caption} 
\usepackage{cuted} 
\usepackage[table]{xcolor}
\usepackage{pifont}
\usepackage{bbding}
\usepackage{newfloat}
\usepackage{listings}
\DeclareCaptionStyle{ruled}{labelfont=normalfont,labelsep=colon,strut=off} 
\floatstyle{ruled}
\newfloat{listing}{tb}{lst}{}
\floatname{listing}{Listing}
\title{Walking Further: Semantic-aware Multimodal Gait Recognition \\ Under Long-Range Conditions}
\author{
    Zhiyang Lu\textsuperscript{\rm 1},
    Wen Jiang\textsuperscript{\rm 1},
    Tianren Wu\textsuperscript{\rm 1},
    Zhichao Wang\textsuperscript{\rm 1}, \\
    Changwang Zhang\textsuperscript{\rm 2},
    Siqi Shen\textsuperscript{\rm 1},
    Ming Cheng\textsuperscript{\rm 1\Envelope}
}
\affiliations{
    \textsuperscript{\rm 1}Fujian Key Laboratory of Sensing and Computing for Smart Cities, Xiamen University \\
    \textsuperscript{\rm 2}OPPO Research Institute\\
    chm99@xmu.edu.cn, changwangzhang@foxmail.com
}

\usepackage{bibentry}

\begin{document}

\maketitle

\begin{abstract}
Gait recognition is an emerging biometric technology that enables non-intrusive and hard-to-spoof human identification. However, most existing methods are confined to short-range, unimodal settings and fail to generalize to long-range and cross-distance scenarios under real-world conditions. To address this gap, we present \textbf{LRGait}, the first LiDAR-Camera multimodal benchmark designed for robust long-range gait recognition across diverse outdoor distances and environments. We further propose \textbf{EMGaitNet}, an end-to-end framework tailored for long-range multimodal gait recognition. To bridge the modality gap between RGB images and point clouds, we introduce a semantic-guided fusion pipeline. A CLIP-based Semantic Mining (SeMi) module first extracts human body-part-aware semantic cues, which are then employed to align 2D and 3D features via a Semantic-Guided Alignment (SGA) module within a unified embedding space. A Symmetric Cross-Attention Fusion (SCAF) module hierarchically integrates visual contours and 3D geometric features, and a Spatio-Temporal (ST) module captures global gait dynamics. Extensive experiments on various gait datasets validate the effectiveness of our method.
\end{abstract}

\begin{links}
    \link{Code}{https://github.com/O-VIGIA/LRGait.git}
\end{links}

\section{Introduction}
In practical applications such as intelligent surveillance and remote identity verification, gait recognition has emerged as a promising biometric technique owing to its non-intrusive nature and robustness over long distances\cite{fan2023opengait-gaitbase,fan2025opengait-tpami,sepas2022deep-gait-survey,zhu2021gait-grew,zheng2022gait-gait3d}. Recent advances have demonstrated strong performance in controlled environments. \textit{However, they are limited to short-range and unimodal settings, and are unexplored under long-range and multimodal conditions.}

\begin{figure*}[ht]
\centering
\includegraphics[width=0.9\textwidth]{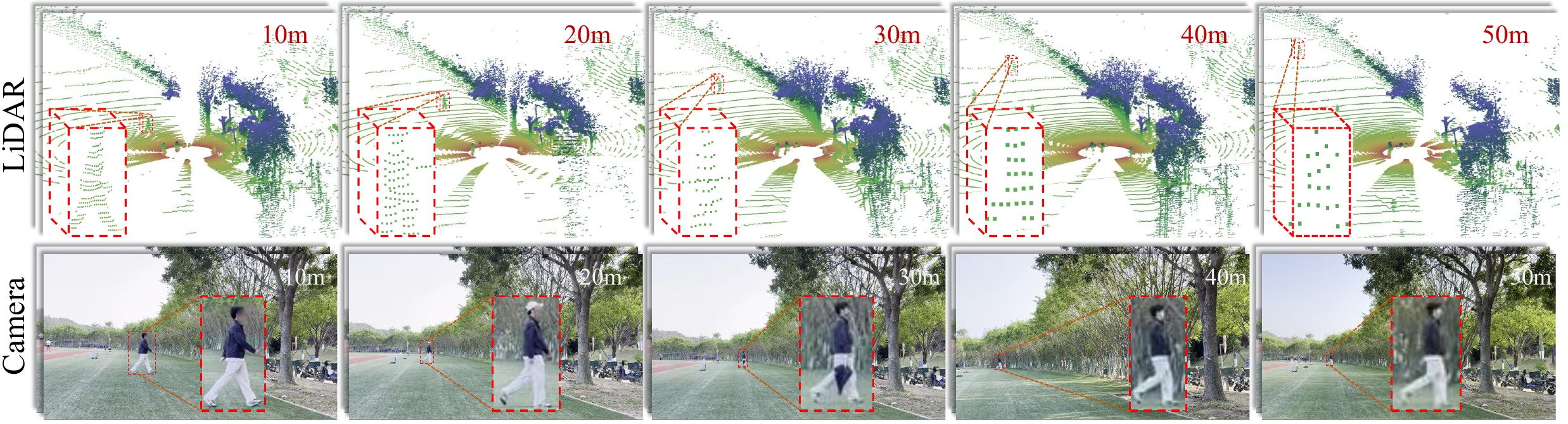}
\caption{Visualization of our proposed multimodal dataset LRGait at various and long-range distances.}
\label{fig::first}
\end{figure*}

Most publicly available gait datasets—such as CASIA-Series~\cite{yu2006framework-casia-b,tan2006efficient-casia-c,song2022casia-e}, OU-MVLP~\cite{takemura2018multi-ou-mvlp}, and Gait3D~\cite{zheng2022gait-gait3d}—are predominantly composed of RGB videos collected within 15 meters, thereby limiting their applicability to real-world surveillance scenarios that demand long-range, cross-distance recognition. Recently, SUSTech1K~\cite{shen2023lidargait-sustech1k} introduced a large-scale LiDAR-Camera multimodal benchmark, laying the groundwork for multimodal gait analysis. FreeGait~\cite{han2024freegait-hmrnet} further advances this direction by capturing gait data in unconstrained outdoor environments. However, SUSTech1K is confined to ranges below 12 meters, and FreeGait includes samples at most 25 meters, underscoring the need for datasets enabling long-range (e.g., 50m) multimodal recognition. Moreover, these datasets lack cross-distance samples per identity, which impedes evaluation under cross-distance cross-view scenarios (e.g.,  50m$\rightarrow$10m). To address these limitations, we introduce LRGait, a long-range cross-distance multimodal gait dataset. It captures synchronized RGB and LiDAR gait sequences across five scopes and eight viewpoints (see Fig.~\ref{fig::first}), encompassing variations in illumination, weather, and carried objects. A comparative overview of existing gait datasets is provided in Table\ref{tab::dataset_compare}. 

Despite recent progress, effectively harnessing the complementary strengths of LiDAR and RGB modalities remains a significant challenge due to the intrinsic modality gap~\cite{bai2025chat-cross-modal-align,chen2025cross-cross-modal-align,li2025alignmamba-cross-modal-align,fan2025opengait-tpami}. Although multimodal gait datasets have emerged, most approaches still rely on unimodal pipelines~\cite{shen2023lidargait-sustech1k,han2024freegait-hmrnet,sepas2022deep-gait-survey}. LiCAF~\cite{deng2024licaf} introduces cross-attention for asymmetric fusion, yet its straightforward fusion strategy struggles to bridge the modality discrepancy. Furthermore, current methods~\cite{han2024freegait-hmrnet,shen2023lidargait-sustech1k,cui2023multi-mmgaitformer} typically adopt depth maps from point clouds or RGB-based silhouettes as pretreatment inputs, resulting in the loss of fine-grained details. \textit{This limitation is particularly exacerbated in long-range or nighttime scenarios, where sparse point clouds and blurred RGB images offer degenerate solutions during pretreatment.}

To address these challenges, we propose EMGaitNet, an end-to-end semantic-guided framework that directly exploits raw RGB videos and point cloud sequences for multimodal long-range gait recognition
. Specifically, we propose a CLIP-based Semantic Mining (SeMi) module that extracts body-part-aware semantic cues. Furthermore, a Semantic-Guided Alignment (SGA) module is designed to reconstruct and align 2D3D cross-modal features by leveraging semantic features. Moreover, a Symmetric Cross-Attention Fusion (SCAF) module is devised to integrate 2D3D features through the cross-attention module hierarchically. Finally, a Spatio-Temporal (ST) module is employed to aggregate global gait dynamics across both spatial and temporal domains. Our contributions are as follows:

\begin{itemize}
    \item We introduce \textbf{LRGait}, the first multimodal gait recognition dataset explicitly designed for long-range and cross-distance scenarios. It comprises synchronized LiDAR and RGB data captured at 5 scopes ($\text{10}m\sim\text{50}m$) from 8 viewpoints. To simulate real-world situations, diverse appearances (e.g., carrying, attire) and environmental variables (e.g., illumination, weather) are incorporated.
    
    \item We propose \textbf{EMGaitNet}, an end-to-end framework that directly inputs raw RGB videos and point cloud sequences for multimodal gait recognition. EMGaitNet incorporates a CLIP-based Semantic Mining (SeMi) module to extract body-part-aware semantics, which guide cross-modal feature alignment in a Semantic-Guided Alignment (SGA) module. Furthermore, we design a Symmetric Cross-Attention Fusion (SCAF) module for deep hierarchical fusion, and a Spatio-Temporal (ST) module to capture global gait dynamics.
\end{itemize}

\section{Related Works}
\subsection{Gait Recognition Methods}
Gait recognition methods are broadly categorized into 2D and 3D representations. 2D appearance-based models~\cite{chao2019gaitset, fan2020gaitpart, huang2021context-cstl, liang2022gaitedge} depend on silhouettes but are sensitive to segmentation quality and appearance changes. Model-based methods~\cite{li2021end, liao2020model, teepe2021gaitgraph} capture gait via pose but suffer from pose estimation errors. Both paradigms lack 3D geometric cues critical for robustness in unconstrained settings. To address this, 3D-based methods have been proposed. Some reconstruct 3D meshes from RGB~\cite{zheng2022gait-gait3d}, while others use LiDAR-derived depth or range images~\cite{shen2023lidargait-sustech1k, ahn20222v}, often losing geometric details during projection. Recent multimodal approaches~\cite{deng2024licaf, cui2023multi-mmgaitformer} integrate 2D and 3D cues for improved performance, yet rely on pre-processed inputs, limiting end-to-end learning. In contrast, our EMGaitNet directly processes raw RGB and LiDAR data, leveraging semantic guidance for end-to-end cross-modal alignment and fusion.

\subsection{Gait Recognition Benchmark}
Gait datasets can be broadly categorized into three types: in-the-lab~\cite{yu2006framework-casia-b, tan2006efficient-casia-c, song2022casia-e, iwama2012isir-ouisir, takemura2018multi, shen2023lidargait-sustech1k}, synthetic~\cite{dou2021versatilegait}, and outdoor/in-the-wild~\cite{mu2021resgait,zheng2022gait-gait3d,han2024freegait-hmrnet}. In-the-lab datasets like the CASIA series support controlled evaluation but are limited to short ranges ($\le$10m) and RGB modality, restricting their utility for data-driven multimodal models. SUSTech1K~\cite{shen2023lidargait-sustech1k} addresses some of these issues by incorporating LiDAR and offering large-scale, multi-view, multimodal data, yet it remains constrained to 12m. Synthetic datasets ease annotation costs but face domain gaps and limited real-world applicability. Outdoor datasets such as FreeGait\cite{han2024freegait-hmrnet} enable evaluation in unconstrained conditions, yet their maximum range remains below 25m. We argue that gait recognition is theoretically feasible at ranges beyond 50m. To this end, we introduce LRGait, a long-range, multimodal dataset designed to explore and push the boundaries of distance-aware gait recognition.

\begin{table*}[htbp] \small
\centering
\begin{tabular}{c|c|c|c|cccc}
\toprule
Dataset & Sensor & Viewpoint & Distance & Outdoor & LR & CD & D\&N \\
\midrule
CASIA-B~\cite{yu2006framework-casia-b} & Camera & 11 & 2m $ \sim $ 4m & \faTimes & \faTimes & \faTimes & \faTimes \\
CASIA-C~\cite{tan2006efficient-casia-c} & Camera & 1 & N/A & \faCheck & \faTimes & \faTimes & \faTimes \\
TUM-GAID~\cite{hofmann2014tum-tum-gaid} & RGB-D & 1 & 3.6m & \faTimes & \faTimes & \faTimes & \faTimes \\
SZTAKI-LGA~\cite{benedek2016lidar-sztaki-lga} & LiDAR & 1 & N/A & \faCheck & \faTimes & \faTimes & \faTimes \\
OU-MVLP~\cite{takemura2018multi-ou-mvlp} & Camera & 14 & 8m & \faTimes & \faTimes & \faTimes & \faTimes \\
GREW~\cite{zhu2021gait-grew} & Camera & 882 & N/A & \faCheck & \faTimes & \faTimes & \faTimes \\
Gait3D~\cite{zheng2022gait-gait3d} & Camera & 39 & N/A & \faCheck & \faTimes & \faTimes & \faTimes \\
CASIA-E~\cite{song2022casia-e} & Camera  & 26 & 8m & \faTimes & \faTimes & \faTimes & \faTimes \\
CCPG~\cite{li2023depth-ccpg} & Camera  & 10 & N/A & \faTimes & \faTimes & \faTimes & \faTimes \\
SUSTech1K~\cite{shen2023lidargait-sustech1k} & LiDAR\&Camera  & 12 & 8m $ \sim $ 12m & \faTimes & \faTimes & \faTimes & \faCheck \\
FreeGait~\cite{han2024freegait-hmrnet} & LiDAR\&Camera  & 3 & 25m & \faCheck & \faTimes & \faTimes & \faCheck \\
\midrule
LRGait(Ours) & LiDAR\&Camera & 8 & 10/20/30/40/50m & \faCheck & \faCheck & \faCheck & \faCheck \\
\bottomrule
\end{tabular}
\caption{Comparison with public datasets for gait recognition, where ``LR" refers to the long-range distance exceeding 30m, ``CD" denotes cross-distance retrieval, and ``D\&N" represents day and night during data collection.} 
\label{tab::dataset_compare}
\end{table*}

\section{Long-Range Gait Benchmark}
\subsection{Overall}
The LRGait dataset is collected using a mobile robot equipped with a 128-beam LiDAR and a monocular RGB camera, capturing synchronized multimodal data. It comprises 5,280 gait sequences from 101 subjects (79 males and 22 females), totaling over 209,000 frames of raw point clouds and RGB images, along with corresponding depth maps and silhouettes. Each subject is recorded walking at distances ranging from 10m to 50m, enabling the study of gait recognition across varying ranges. To simulate various illuminations, 31 subjects were recorded under both day and night environments. To ensure ethical data collection, all participants provided informed consent, and facial regions in RGB images were anonymized via blurring. 

\subsection{Data Collection}
\begin{figure}[t]
\centering
\includegraphics[width=\linewidth]{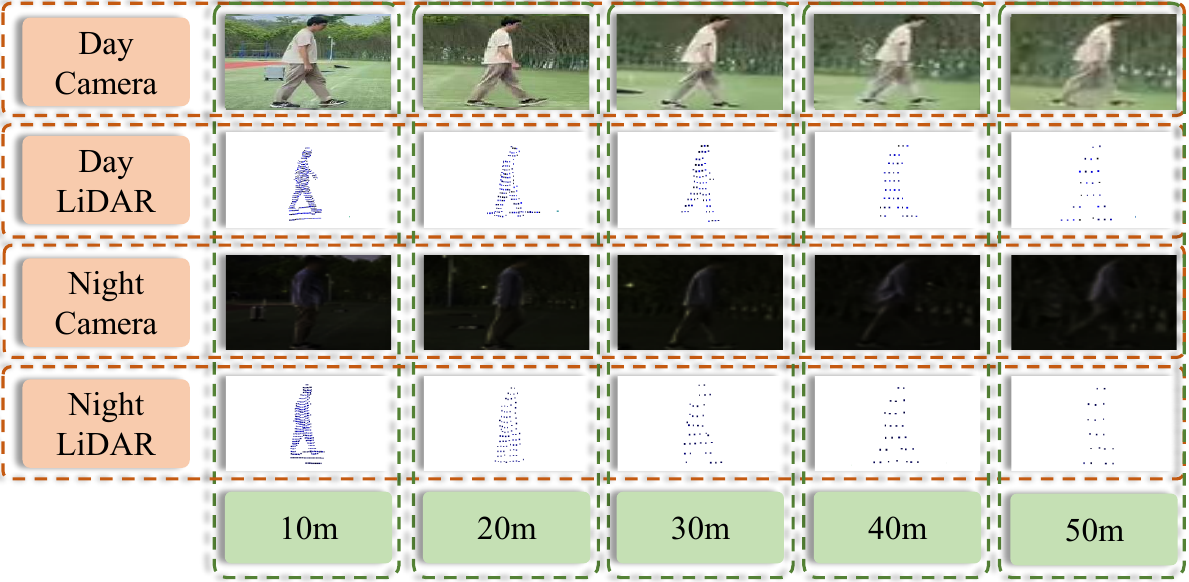}
\caption{Visualizations under daytime and nighttime.}
\label{supfig::day-night}
\end{figure}

The LRGait dataset was collected over a four-week period across three diverse outdoor scenes, covering four distinct weather conditions: sunny, cloudy, overcast, and rainy. We employed an industrial-grade RGB camera and a 128-beam Ouster LiDAR sensor to capture synchronized video and point cloud sequences at 30Hz and 10Hz, respectively. The modalities were temporally aligned during post-processing to ensure accurate cross-modal correspondence. During data collection, each subject was instructed to walk within five various scopes (10, 20, 30, 40, and 50m). To mimic real-world variations, participants were randomly assigned to carry props such as suitcases, umbrellas, and hats. Additionally, for a subset of participants, gait data were recorded under both daytime and nighttime conditions to support research in all-day gait recognition, as shown in Fig.~\ref{supfig::day-night}. To enable multi-view analysis, gait sequences were captured from eight different viewpoints for each walking distance, facilitating robust cross-view and cross-distance gait recognition.

\begin{figure}[t]
\centering
\includegraphics[width=\linewidth]{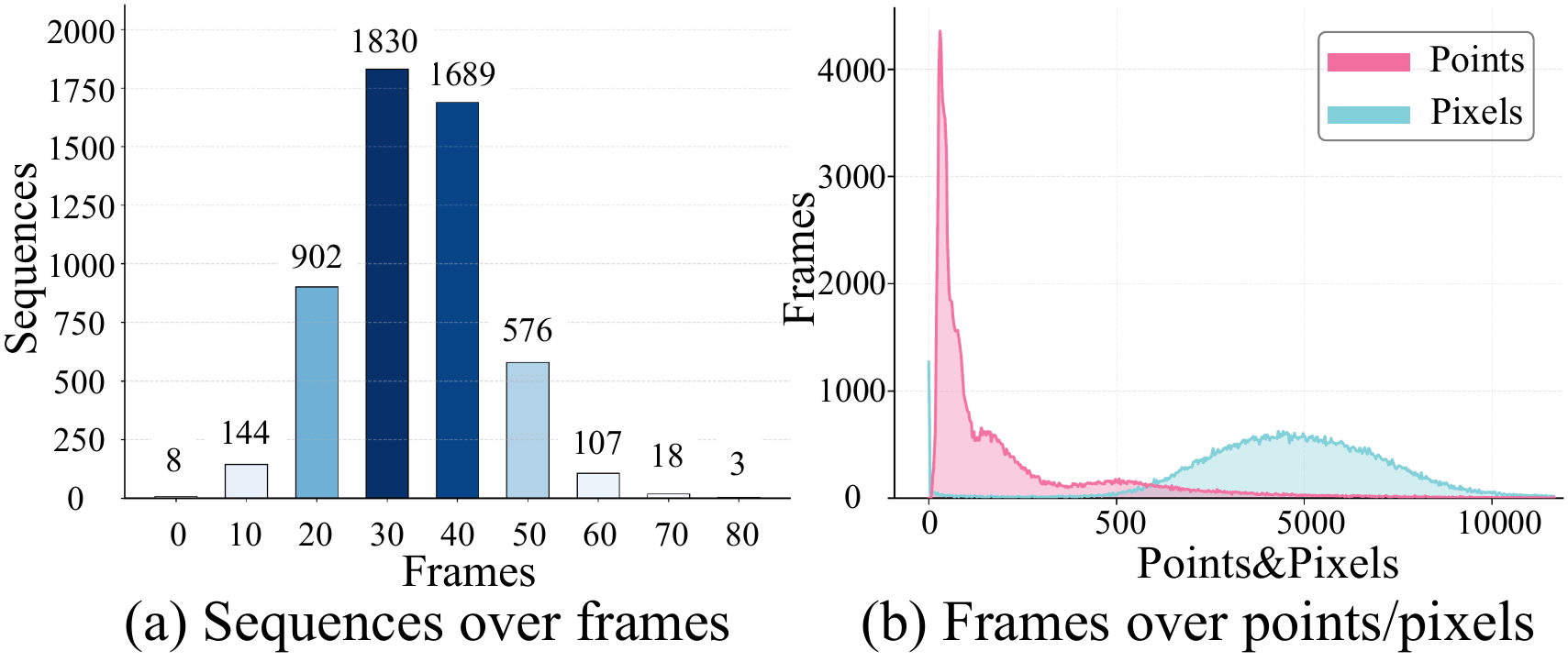}
\caption{Statistics of the proposed LRGait.}
\label{fig::statistic}
\end{figure}

\subsection{Annotations and Representations}
For the camera-based data, we first employed 2D object detection\cite{ge2021yolox} and tracking\cite{zhang2022bytetrack} models to extract gait sequences, followed by silhouette generation using a segmentation model\cite{xie2021segformer}. In challenging conditions such as nighttime or at long distances ($\ge$40m), where the targets are small and visually degraded, we manually annotated 500 frames to fine-tune the detector. For the LiDAR-based data, existing 3D detection frameworks struggle to localize pedestrians reliably at long ranges due to the sparsity of the point cloud. To overcome this limitation, we manually labeled 4,500 frames with pedestrian bounding boxes across various scenes and distances, and used them to train a 3D detection model\cite{lang2019pointpillars}  tailored to long-range scenarios. To address inaccuracies such as false positives and missed detections, we manually corrected the corresponding frames. Upon completion, the entire dataset was thoroughly verified by three expert annotators.

\begin{figure*}[t]
\centering
\includegraphics[width=0.9\textwidth]{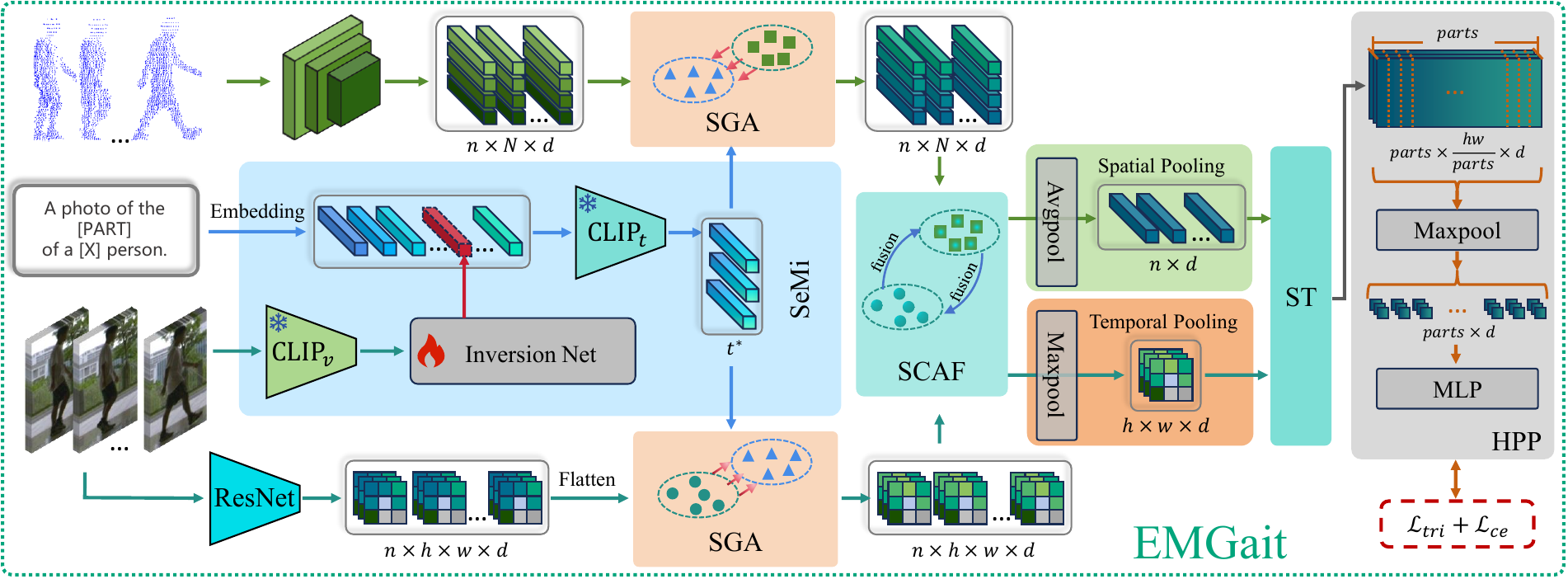}
\caption{Illustration of the proposed framework.}
\label{fig::framework}
\end{figure*}

\subsection{Statistics and Evaluation Metrics}

Fig.~\ref{fig::statistic} illustrates the data distribution of the LRGait dataset. The evaluation protocol adopts the cross-view recognition setting, following the standard used in SUSTech1K\cite{shen2023lidargait-sustech1k}, in which probe features are matched against gallery features across various views. Probe sets are partitioned by distance to assess attribute influence in cross-view retrieval. Rank-$1$ and Rank-$5$ accuracies are used as evaluation metrics. Refer to the supplementary materials for statistical details.

\section{Semantic-Guided Multimodal Gait Recognition}
\subsection{Problem Definition}
We design an end-to-end multimodal gait recognition framework that directly takes RGB video
\begin{equation}
    I=\left \{ I_{i}^{j}|i=1,2,\dots ,m;j=1,2,\dots,n \right \} 
\end{equation}
and point cloud sequences
\begin{equation}
    P=\left \{ P_{i}^{j}|i=1,2,\dots ,m;j=1,2,\dots,n\right \} 
\end{equation}
as input, where $m$ is the number of gait sequences and $n$ is the number of frames per sequence. Specifically, $I_{i}^{j} \in \mathbb{R}^{H \times W \times 3}$ denotes an RGB frame with height $H$ and width $W$, and $P_{i}^{j} \in \mathbb{R}^{N \times 3}$ is the corresponding LiDAR point cloud containing $N$ number of points.
Our goal is to learn discriminative multimodal gait representations directly from the raw RGB and point cloud data:
\begin{equation}
    F=G_{\theta }\left ( I,P \right ) ,
\end{equation}
where $F=\left \{ F_{i}|i=1,2,\dots ,m \right \} $ denotes the learned gait features, and $G_{\theta }$ is our proposed multimodal framework.

\subsection{Feature Extraction}
For the camera modality, we adopt the lightweight ResNet9 backbone from OpenGait\cite{fan2023opengait-gaitbase} to extract frame-wise visual features, formulated as:
\begin{equation}
    F^{2d}_{i,j}=\mathrm{ResNet}_{9}\left (  I_{i}^{j}\right ),
\end{equation}
where $F^{2d}_{i,j} \in \mathbb{R}^{h\times w\times d} $ denotes the spatial feature map for the $j$-th frame in the $i$-th sequence, with $h$, $w$, and $d$ being the height, width, and channel dimensions, respectively. For the LiDAR modality, we adopt a PointGNN-based backbone\cite{shi2020point-pointgnn} to extract discriminative 3D features, which mitigates the adverse effects of point cloud sparsity. 
Specifically, given a point cloud sequence $P_{i}^{j}$ and its corresponding feature representation $F^{3d}_{i,j}$ (initialized as coordinates), we first construct a local neighborhood for each point:
\begin{equation}
    A_{i,j}[k]=\left \{  P_{i}^{j}[k],\mathcal{N}_{P_{i}^{j}}\left ( P_{i}^{j}[k] \right )   \right \} ,
\end{equation}
where $k\in\left \{ 1,2,\dots,N \right \} $, $P_{i}^{j}[k]$ denotes the $k$-th point, and $\mathcal{N}_{P_{i}^{j}}\left ( P_{i}^{j}[k] \right ) $ retrieves its spatial neighbors in $P_{i}^{j}$. The adjacency relationship is defined based on the cosine similarity between the feature vectors of points, defined as:
\begin{equation}
    \mathrm{cos}(F_{i,j}^{3d}[k], F_{i,j}^{3d}[u]) = \frac{F_{i,j}^{3d}[k] \cdot F_{i,j}^{3d}[u]}{\|F_{i,j}^{3d}[k]\|_2 \cdot \|F_{i,j}^{3d}[u]\|_2},
\end{equation}
where $F_{i,j}^{3d}[k] \in \mathbb{R}^{d}$ denotes the feature vector of $P_{i}^{j}[k]$. Using the similarity scores, we construct a local graph by selecting the TopK most similar points from the entire point set as neighbors:
\begin{equation}
    \mathcal{N}_{P_{i}^{j}}(P_{i}^{j}[k]) = \underset{u \neq k}{\mathrm{TopK}} \left( \mathrm{cos}(F_{i,j}^{3d}[k], F_{i,j}^{3d}[u]) \right).
\end{equation}
We compute the edge features in the local graph as:
\begin{equation}
    e_{k,u}=\mathrm{Concat}\left [ P_{i}^{j}[u]-P_{i}^{j}[k],F_{i,j}^{3d}[k], F_{i,j}^{3d}[u] \right ] .
\end{equation}
Here, $\mathrm{Concat}$ denotes the concatenation operation along the feature dimension. Subsequently, a multi-layer perceptron (MLP) layer is applied to introduce nonlinearity, followed by feature aggregation within each local neighborhood to update the point features from the previous layer:
\begin{equation}
    F_{i,j}^{3d}[k]=\mathop{\mathrm{ Maxpool}} \limits_{u\in\mathcal{N}_{P_{i}^{j}}\left ( P_{i}^{j}[k] \right ) }\left ( \mathrm{MLP}\left ( e_{k,u} \right )   \right ) .
\end{equation}
By stacking multiple such graph convolution layers, the model progressively captures both local and global geometric patterns, yielding the final representation $F^{3d}_{i,j} \in \mathbb{R}^{N \times d}$. Our overall framework is illustrated in Fig.~\ref{fig::framework}.

\subsection{CLIP-Based Semantic Mining}
The modality gap between 2D images and 3D point clouds frequently results in suboptimal fusion during feature integration. Moreover, background noise introduces ineffective fusion, exemplified by the blending of LiDAR point clouds with irrelevant RGB background areas. To address this, we propose a CLIP-based Semantic Mining (SeMi) module that leverages semantic cues to align cross-modal features and enhance regional correspondence. Specifically, we construct explicit body-part prompts and feed them into the CLIP text encoder $\mathrm{CLIP}_t$ to obtain semantic cues. A general template—``A photo of the [PART] of a [X] person"—is instantiated with body-part terms, using a predefined list [``head", ``arms", ``torso", ``legs", ``feet"] to generate fine-grained semantic descriptions, which are then tokenized into embeddings $t \in \mathbb{R}^{5 \times l \times d}$. However, these class-level semantics fall short in capturing instance-level nuances for gait recognition. Inspired by PromptSG\cite{yang2024pedestrian-PromptSG}, we replace [X] with visual embeddings derived from the inversion net to generate identity-aware descriptions. Specifically, for an input image $I_i^j$, we utilize CLIP’s visual encoder to extract the global visual embedding:
\begin{equation}
    v=\mathrm{CLIP}_{v}\left ( I_{i}^{j} \right ),
\end{equation}
where $v\in\mathbb{R}^{1\times d}.$
We then employ an inversion network to map the visual feature $v$ from visual space into the text space, formulated as: $v^{*}  = F_{inv}\left ( v \right ).$ The transformed feature $v^{*}$ is leveraged as a pseudo token to replace the [X] placeholder in the prompt for fine-grained and identity-aware semantic token embedding. The modified prompts are then fed into the text encoder $\mathrm{CLIP}_t$ to extract body-part-aware semantic features, denoted as $t^{*} \in \mathbb{R}^{5 \times d}$.

\begin{figure}[t]
\centering
\includegraphics[width=\linewidth]{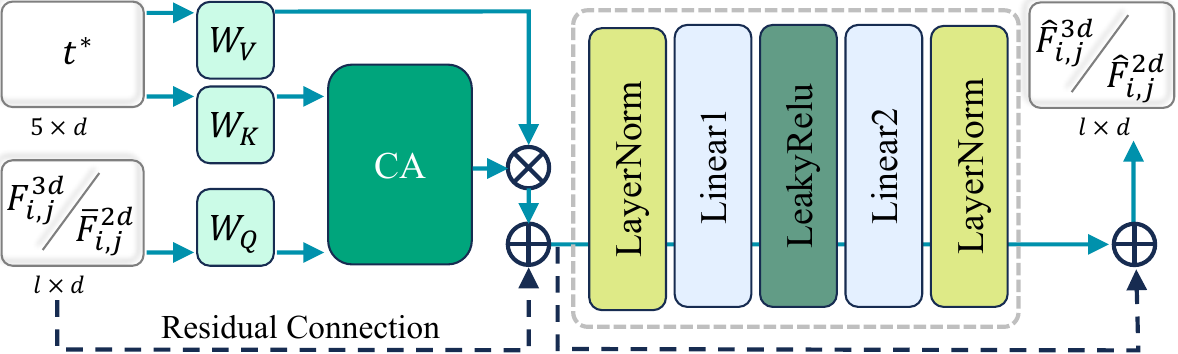}
\caption{Details of the proposed SGA module.}
\label{fig::SGA}
\end{figure}

\subsection{Semantic-Guided Alignment Module}
We leverage multi-grained semantic cues as an intermediate bridge to mitigate the modality gap between RGB images and point cloud data, while simultaneously suppressing feature noise, as shown in Fig.~\ref{fig::SGA}. Specifically, we design the feature alignment module based on a cross-attention mechanism. Given a 2D feature map $F^{2d}_{i,j}$, we flatten its spatial dimensions to obtain $\bar{F}^{2d}_{i,j}\in \mathbb{R}^{hw\times d} $ that match the format of the 3D feature $F^{3d}_{i,j}$. Taking $\bar{F}^{2d}_{i,j}$ as the query and the semantic features $t^*$ as both key and value, we perform cross-attention fusion, formally defined as:
\begin{equation}
    \mathrm{CA}\left ( \bar{F}^{2d}_{i,j},t^{*}   \right ) =\mathrm{Softmax}\left ( \frac{QK^{\top}}{\sqrt{d} }  \right )V,
\end{equation}
where $Q = \bar{F}^{2d}_{i,j}W_Q,K = t^{*} W_K,V = t^{*} W_V$, with $W_Q, W_K, W_V \in \mathbb{R}^{d \times d}$ denoting learnable projection matrices. Furthermore, we integrate residual connections and a Feed Forward Network (FFN) following the attention output:
\begin{equation}
    \tilde{F}^{2d}_{i,j} = \mathrm{LayerNorm}\left( \bar{F}^{2d}_{i,j} + \text{CA}(\bar{F}^{2d}_{i,j}, t^{*}) \right),
\end{equation}
\begin{equation}
    \hat{F}^{2d}_{i,j} = \mathrm{LayerNorm}\left( \tilde{F}^{2d}_{i,j} + \text{FFN}(\tilde{F}^{2d}_{i,j}) \right).
\end{equation}
Here, $\mathrm{FFN}(\cdot)$ consists of two MLP layers with LeakyReLU activation, and the refined 2D features are denoted by $\hat{F}^{2d}_{i,j} \in \mathbb{R}^{hw \times d}$. Similarly, we employ semantic features $t^*$ to align the point cloud features, yielding the refined $\hat{F}^{3d}_{i,j} \in \mathbb{R}^{N \times d}$.

\begin{figure}[t]
\centering
\includegraphics[width=\linewidth]{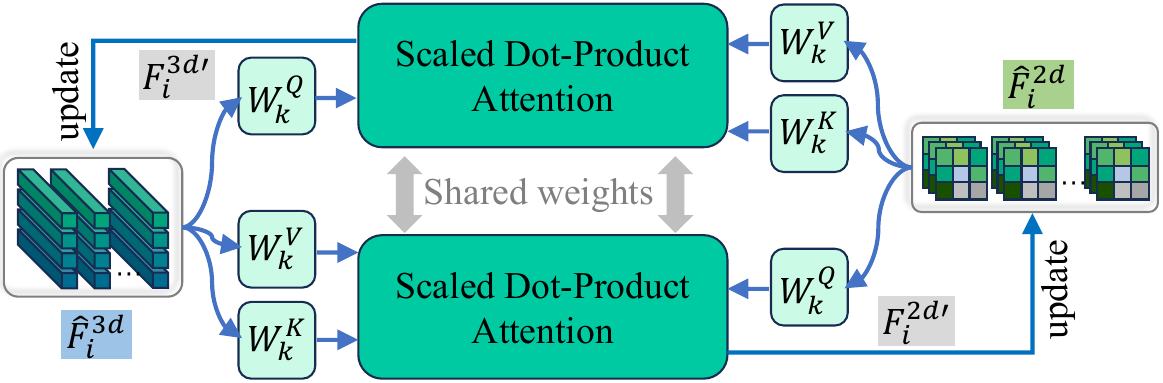}
\caption{Details of the proposed SCAF module.}
\label{fig::SCAF}
\end{figure}

\subsection{Symmetric Cross-Attention Fusion Module}
After cross-modal alignment, the features from different modalities are projected into a shared latent space, where a symmetric multi-head cross-attention is applied to integrate complementary information, as presented in Fig.~\ref{fig::SCAF}. Specifically, we construct a symmetric dual-stream module, where the refined image features $\hat{F}^{2d}_{i,j}$ and point cloud features $\hat{F}^{3d}_{i,j}$ alternately serve as query and attend to each other as key and value, enabling bidirectional alignment and mutual information fusion. For each fusion layer, the attention-based update of the image stream is computed as:
\begin{equation}
    F^{2d'}_{i,j} = \mathrm{Concat}(\text{head}_1, \ldots, \text{head}_h) W^O,
\end{equation}
where $h$ denotes the number of attention heads, and $W^O \in \mathbb{R}^{h d_h \times d}$ represents the output projection matrix. The operation for each attention head is defined as follows:
\begin{equation}
    \text{head}_k = \mathrm{Softmax}\left( \frac{Q_k K_k^\top}{\sqrt{d_h}} \right) V_k,
\end{equation}
\begin{equation}
    Q_k = F^{2d}_{i,j} W^Q_k,\quad 
K_k = F^{3d}_{i,j} W^K_k,\quad 
V_k = F^{3d}_{i,j} W^V_k,
\end{equation}
where $W^Q_k, W^K_k, W^V_k \in \mathbb{R}^{d \times d_{h}}$ are learnable projection matrices. In parallel, the point cloud stream is updated analogously with shared weights, yielding $F^{3d'}_{i,j}$. The refined features are subsequently propagated to the next stage of hierarchical fusion by updating $\hat{F}^{2d}_{i,j} = F^{2d'}_{i,j}$ and $\hat{F}^{3d}_{i,j} = F^{3d'}_{i,j}$.

\subsection{Spatial Temporal Module}
To capture temporal dynamics, we apply Temporal Pooling over the sequence of the fused sequence features $\hat{F}^{2d}_{i} =\{  \hat{F}^{2d}_{i,j} \}_{j=1}^{n}$, where $\hat{F}^{2d}_{i} \in \mathbb{R} ^{n\times h\times w\times d}$.
Meanwhile, Spatial Pooling is performed on the point cloud feature sequence $\hat{F}^{3d}_{i} =\{  \hat{F}^{3d}_{i,j} \}_{j=1}^{n}$ to obtain global spatial features, where $\hat{F}^{3d}_{i} \in \mathbb{R} ^{n\times N\times d}$. These operations are formulated as:
\begin{equation}
    F_{i}^{tp}=\mathop{\mathrm{Maxpool}} \limits_{n}\left ( \hat{F}^{2d}_{i} \right ),
\end{equation}
\begin{equation}
    F_{i}^{sp}=\mathop{\mathrm{Avgpool}} \limits_{N}\left ( \hat{F}^{3d}_{i} \right ) ,
\end{equation}
where $F_{i}^{tp} \in \mathbb{R}^{h \times w \times d}$ represents the aggregated temporal feature, and $F_{i}^{sp} \in \mathbb{R}^{n \times d}$ denotes the global spatial feature. We leverage the $\mathrm{CA}$ to integrate the spatio-temporal features:
\begin{equation}
    \tilde{F}_{i}^{sp}=\mathrm{CA}\left ( F_{i}^{sp},F_{i}^{tp} \right )  +F_{i}^{sp},
\end{equation}
\begin{equation}
    F_{i}^{fusion}=\mathrm{MLP}\left (  \mathrm{CA}\left ( F_{i}^{tp}, \tilde{F}_{i}^{sp}\right )+F_{i}^{tp}\right ) ,
\end{equation}
where $F_{i}^{fusion} \in \mathbb{R} ^{h\times w\times d}$. Subsequently, the Horizontal Pyramid Pooling (HPP) module\cite{fan2025opengait-tpami} is employed to perform part-based matching.

\section{Training and Inference}
Following prior works\cite{fan2023opengait-gaitbase,fan2025opengait-tpami},  we optimize the model using a combination of triplet loss and cross-entropy loss. The objective functions are formulated as:
\begin{equation}
    \mathcal{L}=\alpha  \mathcal{L}_{tri}+\beta \mathcal{L}_{ce},
\end{equation}
where $\alpha=1.0,\beta=2.0$ in default. 
During inference, the L2 Euclidean distance between features is used to measure the similarity between the probe and gallery samples.

\section{Experiments}
\begin{table*}[t] \small

\centering
\begin{tabular}{c|c|cccccccc|cc}
\toprule
\multirow{2}{*}{Methods} & \multirow{2}{*}{Modality} & \multicolumn{8}{c|}{Probe Sequence (Rank-$1$ Accuracy \%)}  & \multicolumn{2}{c}{Overall} \\
\cmidrule(r){3-12}
& & NM & BG & CL & CR & UB & UN & OC & NT & R-$1$ & R-$5$ \\
\midrule
GaitSet\cite{chao2019gaitset} & sil & 69.1 & 68.3 & 37.4 & 65.0 & 63.1 & 67.2 & 61.0 & 23.0 & 65.0 & 84.8 \\
GaitBase\cite{fan2023opengait-gaitbase} & sil & 81.3 & 77.3 & 49.6 & 75.7 & 75.4 & 76.7 & 81.4 & 25.8 & 76.0 & 89.1 \\
\midrule
SimpleView\cite{goyal2021revisiting-simpleview} & depth & 72.3 & 68.8 & 57.2 & 63.3 & 49.2 & 79.7 & 62.5 & 66.5 & 64.8 & 85.8\\
LidarGait\cite{shen2023lidargait-sustech1k} & depth & 91.8 & 88.6 & 74.6 & 89.0 & 67.5 & 80.9 & 94.5 & \underline{90.4} & 86.8 & 96.1 \\

PointTransformer\cite{zhao2021point-pointtransformer} & pc & 53.2 & 48.1 & 32.0 & 43.2 & 39.1 & 47.9 & 41.8 & 47.1 & 44.4 & 76.7 \\

PointNet++\cite{qi2017pointnet++} & pc & 82.5 & 78.7 & 58.7 & 76.1 & 74.0 & 85.4 & 75.8 & 74.8 & 77.1 & 94.1 \\

LidarGait++\cite{shen2025lidargait++} & pc & 94.2 & 93.9 & 79.7 & 92.4 & \underline{91.5} & \textbf{96.6} & 91.9 & \textbf{92.2} & 92.7 & 98.2 \\
\midrule
HMRNet\cite{han2024freegait-hmrnet} & pc+depth & 92.7 & 92.3 & 79.6 & 90.3 & 83.1 & \underline{95.2} & 86.2 & \underline{90.4} & 90.2 & 97.5 \\
\midrule
MMGaitFormer\cite{cui2023multi-mmgaitformer} & depth+sil & 94.3 & 93.7 & 80.0 & 91.8 & 84.0 & 88.7 & 95.7 & 86.0 & 91.1 & 98.2 \\

LiCAF\cite{deng2024licaf} & depth+sil & \underline{95.8} & \underline{95.7} & \textbf{82.7} & \underline{94.5} & 89.3 & 93.6 & \underline{96.6} & 88.7 & \underline{93.9} & \underline{98.8} \\
\midrule
MultiGait++\cite{jin2025exploring-multigait++} & sil+ps+fl & 92.0 & 89.4 & 50.4 & 87.6 & 89.7 & 89.1 & 93.4 & 45.1 & 87.4 & 95.6 \\
\midrule
\rowcolor{myteal!40}
EMGaitNet(Ours) & pc+rgb & \textbf{98.2} & \textbf{96.4} & \underline{81.7} & \textbf{96.2} & \textbf{94.9} & 93.6 & \textbf{99.6} & 88.1 & \textbf{96.0} & \textbf{99.0} \\

\bottomrule
\end{tabular}
\caption{Comparison of cross-view Rank-$1$ accuracy under various conditions on the SUSTech1K dataset. The best results are highlighted in \textbf{bold}, while the second-best entries are \underline{underlined}.} 
\label{tab::comparision_sustech1k}
\end{table*}

\begin{table*}[t] \small
\centering
\begin{tabular}{c|c|cccccccc|cc}
\toprule
\multirow{2}{*}{Methods} & \multirow{2}{*}{Modality} & \multicolumn{8}{c|}{Probe Sequence (Rank-$1$ Accuracy \%)}  & \multicolumn{2}{c}{Overall} \\
\cmidrule(r){3-12}
& & D-20  & D-30  & D-40  & D-50  & N-20  & N-30  & N-40  & N-50  & R-$1$ & R-$5$ \\
\midrule
GaitBase\cite{fan2023opengait-gaitbase} & sil & 67.9 & 53.9 & 48.5 & 33.8 & \underline{41.6} & \textbf{33.4} & \underline{19.8} & \underline{13.2} & 46.8 & 77.8 \\
\midrule
LidarGait\cite{shen2023lidargait-sustech1k} & depth & 26.1 & 14.6 & 13.2 & 10.8 & 18.5 & 14.6 & 10.6 & 9.6 & 15.7 & 51.3 \\
LidarGait++\cite{shen2025lidargait++} & pc & 32.1 & 24.4 & 18.8 & 12.6 & 23.1 & 13.2 & 11.7 & 11.2 & 20.9 & 59.0 \\
\midrule
HMRNet\cite{han2024freegait-hmrnet} & pc+depth & 58.7 & 53.1 & 50.3 & 44.8 & 22.5 & 19.3 & 11.8 & 8.6 & 41.1 & 71.7 \\
\midrule
MMGaitFormer\cite{cui2023multi-mmgaitformer} & depth+sil & 72.4 & 70.2 & 58.1 & 62.7 & 40.3 & 26.8 & 15.5 & 7.6 & 57.1 & 80.4 \\
LiCAF\cite{deng2024licaf} & depth+sil & \underline{74.8} & \underline{71.6} & \underline{60.4} & \underline{65.3} & \textbf{42.5} & 27.8 & 14.6 & 9.9 & \underline{59.6} & \underline{82.9} \\
\midrule
\rowcolor{myteal!40}
EMGaitNet (Ours) & pc+rgb & \textbf{88.5} & \textbf{82.4} & \textbf{80.8} & \textbf{74.4} & 38.2 & \underline{31.7} & \textbf{21.9} & \textbf{17.1} & \textbf{68.9} & \textbf{85.8} \\
\bottomrule
\end{tabular}
\caption{Comparison of cross-distance cross-view Rank-$1$ accuracy under various settings on the proposed LRGait dataset.} 
\label{tab::comparision_lrgait}
\end{table*}




\begin{table}[t] 
\centering
\resizebox{\linewidth}{!}{
\begin{tabular}{c|c|ccc}
\toprule
Methods & Modality & R-$1$ & R-$5$ & mAP \\
\midrule
GaitSet\cite{chao2019gaitset} & sil & 57.1 & 71.9 & 64.0 \\
GaitBase\cite{fan2023opengait-gaitbase} & sil & 62.6 & 75.3 & 68.6 \\
\midrule
LidarGait\cite{shen2023lidargait-sustech1k} & depth & 74.2 & 88.8 & 80.7 \\
PointNet++\cite{qi2017pointnet++} & pc & 59.3 & 81.2 & 69.3 \\
LidarGait++\cite{shen2025lidargait++} & pc & \underline{82.0} & \underline{93.6} & \underline{87.2} \\
\midrule
HMRNet\cite{han2024freegait-hmrnet} & pc+depth & 80.8 & \underline{93.6} & 86.5 \\
\midrule
\rowcolor{gray!20}
EMGaitNet(Ours) & pc+rgb & \textbf{85.2} & \textbf{96.8} & \textbf{89.0} \\

\bottomrule
\end{tabular}
}
\caption{Comparisons with SOTA methods on FreeGait.}
\label{tab::comparsion_freeGait}
\end{table}

\subsection{Implementation Details}
In SUSTech1K\cite{shen2023lidargait-sustech1k}, we apply Farthest Point Sampling (FPS) to downsample each point cloud (pc) frame to 512 points to facilitate end-to-end model input. For LRGait, following the same protocol, RGB images (rgb) and silhouettes (sil) are resized to 64×64. Each point cloud frame is downsampled to 256 points via FPS, and the corresponding depth projections (depth) are also resized to 64×64. In FreeGait, we follow the same preprocessing as HMRNet\cite{han2024freegait-hmrnet}. The model is trained for 40,000 epochs using the Adam optimizer with a weight decay of 0.0005. The initial learning rate is set to 0.0003 for the SUSTech1K and FreeGait datasets, and 0.00005 for the LRGait dataset. A MultiStepLR scheduler is employed to decay the learning rate by a factor of 0.1 at the 15,000th and 30,000th epochs. During each training epoch, 10 RGB frames and their corresponding point cloud frames are randomly sampled for end-to-end training. All experiments are implemented and evaluated on two NVIDIA RTX 3090 GPUs.

\subsection{Results and Analysis}
We compare our approach against advanced unimodal and multimodal methods. The proposed EMGait achieves superior performance across all benchmarks, establishing new state-of-the-art results not only on the long-range and cross-distance LRGait dataset but also on the large-scale, multi-view, and multi-variable SUSTech1K, as well as the highly challenging FreeGait dataset.

\subsubsection{SUSTech1K}
Table~\ref{tab::comparision_sustech1k} provides a detailed comparison of the SUSTech1K dataset. Specifically, the rgb and pc correspond to the raw inputs captured by the camera and LiDAR, respectively. The sil and depth indicate the preprocessed representations obtained from RGB and point cloud inputs\cite{fan2023opengait-gaitbase,shen2025lidargait++}. Meanwhile, ps and fl denote higher-level semantic cues extracted from rgb, representing human parsing and optical flow\cite{jin2025exploring-multigait++}. On the Overall Rank-$1$ metric, our EMGaitNet achieves an accuracy of 96.0\%, establishing a new state-of-the-art (SOTA) by outperforming the second-best method by 2.2\%. This highlights EMGaitNet’s capability to not only capture 2D cues such as human shape and contours, but also extract 3D representations involving body dimensions and skeletal topology. Compared to the previous SOTA unimodal end-to-end approach LidarGait++\cite{shen2025lidargait++}, EMGaitNet improves Overall Rank-$1$ by 3.6\%. Remarkably, it reaches 99.6\% accuracy under occlusion conditions, evidencing its robust understanding of multimodal priors such as body structure and shape. 

\begin{figure}[t]
\centering
\includegraphics[width=\linewidth]{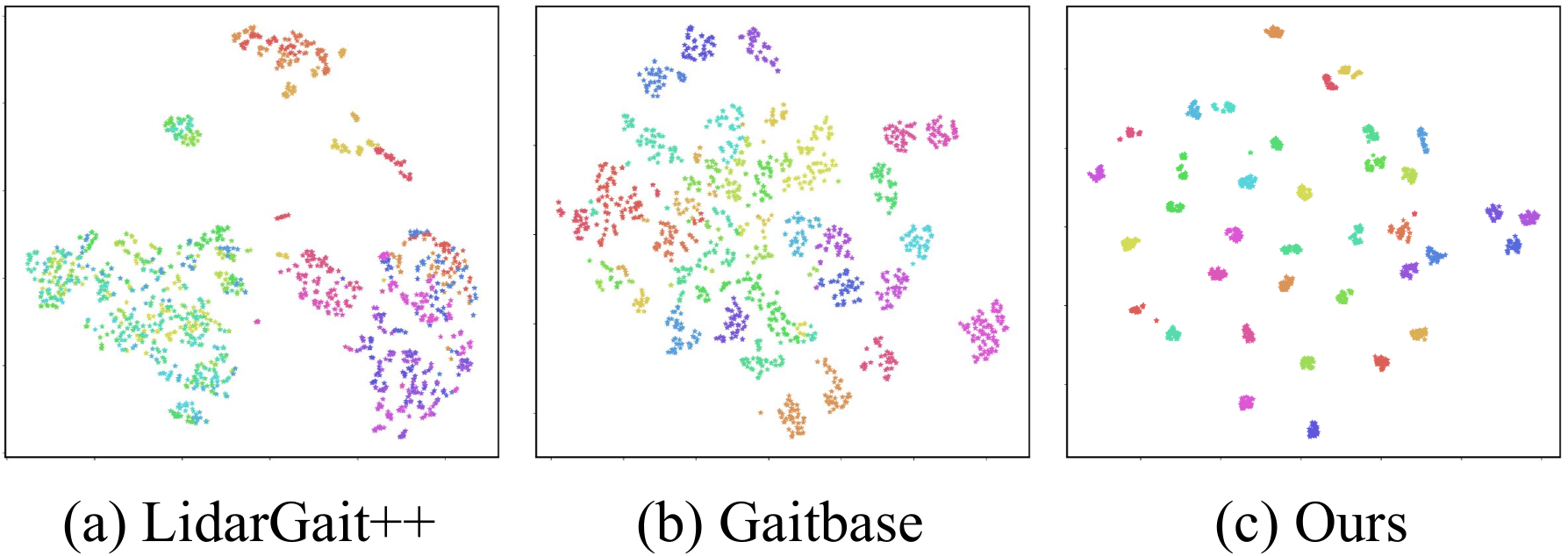}
\caption{Comparison of t-SNE visualizations.}
\label{fig::tsne}
\end{figure}

\subsubsection{LRGait}
We conduct a comparative study of multiple approaches on our proposed LRGait dataset, with the day-10m (D-10) setting designated as the gallery. EMGaitNet delivers SOTA performance overall and cross-distance retrieval under day conditions, as presented in Table~\ref{tab::comparision_lrgait}. Notably, it reaches 74.4\% accuracy at a distance of day-50m (D-50), outperforming the second-best method by a significant margin of 14.0\%. These results affirm the effectiveness of our end-to-end multimodal framework in integrating the 3D geometric structures from point clouds with the semantic cues from RGB images, enabling robust performance across long-range and cross-distance. Fig.~\ref{fig::tsne}
shows advancement of our EMGaitNet in the t-SNE~\cite{maaten2008visualizing-tsne} manner. 
In contrast, under night (N) conditions, the performance of nearly all methods deteriorates markedly, largely due to the severe domain shift between day and night. Consequently, developing multimodal domain-adaptive approaches for cross-distance gait recognition under day-night conditions is a crucial and promising research direction.

\subsubsection{FreeGait}
Table~\ref{tab::comparsion_freeGait}  presents the results on FreeGait. Our EMGaitNet outperforms LiDARGait++\cite{shen2025lidargait++} by 3.9\%, 3.4\%, and 2.1\% on Rank-$1$, Rank-$5$, and mAP, respectively, demonstrating its ability to learn highly discriminative and robust multimodal gait features.

\begin{table}[t] \footnotesize
\centering
\resizebox{\linewidth}{!}{
\begin{tabular}{cccc|cc}
\toprule
Baseline & SeMi & SGA & ST & Overall R-$1$ & Overall R-$5$ \\
\midrule
\checkmark &  &  &  & 52.3 & 70.2 \\
\checkmark &  & \checkmark &  & 58.5 & 75.9 \\
\checkmark & \checkmark & \checkmark &  & 64.2 & 80.7 \\
\checkmark & \checkmark & \checkmark & \checkmark & \textbf{68.9} & \textbf{85.8} \\
\bottomrule
\end{tabular}
}
\caption{Ablation studies on the LRGait dataset.} 
\label{tab::ablation}
\end{table}

\subsection{Ablation Study}
To validate the effectiveness of each component in our proposed EMGaitNet, we conduct comprehensive ablation studies on the LRGait dataset, as shown in Table~\ref{tab::ablation}. Starting from a baseline that directly fuses RGB and point cloud features, we observe a Rank-$1$ accuracy of 52.3\%, indicating the inherent challenge in multimodal fusion under long-range conditions. Introducing the Semantic-Guided Alignment (SGA) module significantly boosts accuracy to 58.5\%, highlighting its critical role in mitigating modality gaps between 2D and 3D features. Further integrating the CLIP-based Semantic Mining (SeMi) module improves performance to 64.2\%, demonstrating the importance of human semantic cues in guiding feature representation. Finally, the inclusion of the Spatio-Temporal (ST) module increases the Rank-$1$ accuracy to 68.9\%, demonstrating the benefit of modeling global gait dynamics.

\section{Conclusion and Future Work}
In this paper, we introduce LRGait, the first large-scale multimodal gait dataset designed for long-range and cross-distance scenarios, along with EMGaitNet, a semantic-guided, end-to-end multimodal fusion framework. LRGait comprises 101 subjects, with gait sequences captured across distances ranging from 10m to 50m, under diverse lighting, weather, and environmental conditions. Compared with existing public gait datasets—none of which exceed 25m in collection range—LRGait significantly expands the distance frontier, with the ambition of advancing effective gait recognition beyond 50m. Our proposed EMGaitNet achieves SOTA performance on LRGait, SUSTech1K, and FreeGait, demonstrating the effectiveness of leveraging both raw RGB videos and LiDAR point clouds in a unified end-to-end pipeline.

While our method performs robustly under daytime conditions on LRGait—achieving over 74\% Rank-$1$ accuracy within the scope of 50m—it still faces challenges in nighttime scenarios, leaving substantial room for future improvement. We hope our contributions inspire further research toward more scalable and resilient multimodal gait recognition systems, ultimately enabling gait analysis to go farther, across longer distances and more diverse conditions.



\section{Supplementary Materials}

\begin{strip}
\centering
\includegraphics[width=\textwidth]{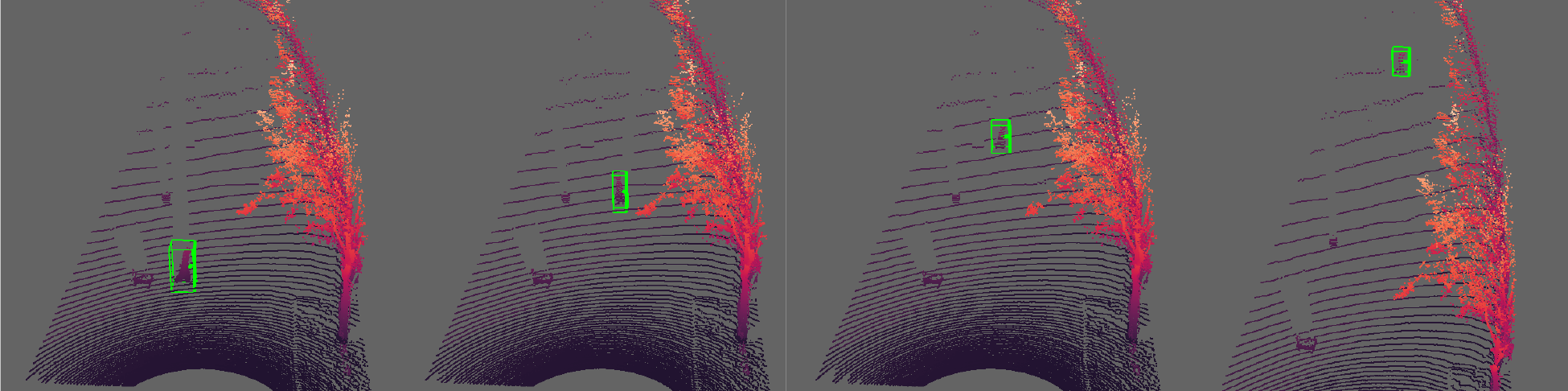}
\captionof{figure}{Point cloud object detection annotation examples.}
\label{supfig::pc1}
\end{strip}

\begin{figure}[t]
\centering
\includegraphics[width=\linewidth]{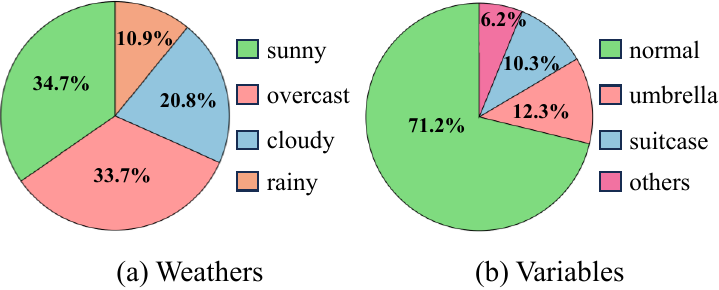}
\caption{Statistical overview of weather conditions and behavioral variations in LRGait. (a) Number of subjects captured under different weather scenarios. (b) Distribution of activity-related variations (e.g., backpacks, suitcases) across all frames in the dataset.}
\label{supfig::weathers_variables}
\end{figure}

\subsection{Long-range Pedestrian Detection Benchmark}
Accurate pedestrian detection from LiDAR point clouds at long ranges remains a significant challenge due to the extreme sparsity and noise in the data, particularly beyond 30 meters. Most existing 3D object detection models, which are typically designed for dense point clouds in autonomous driving scenarios, fail to generalize effectively under such sparse conditions. To facilitate reliable evaluation and provide high-quality supervision for future detection and recognition tasks, we manually constructed a long-range pedestrian detection benchmark tailored to our LRGait dataset.

Specifically, three trained annotators spent approximately 80 hours labeling a total of 4,500 LiDAR frames. Each frame was carefully examined to ensure accurate bounding box annotations of visible pedestrians at varying distances. This annotation effort not only supports further research in long-range pedestrian detection but also provides valuable ground-truth data for evaluating multimodal fusion frameworks.

Examples of the annotated point cloud frames are presented in Fig.~\ref{supfig::pc1}, which demonstrate the difficulty of long-range pedestrian detection and highlight the necessity of high-quality annotations for benchmarking future models in realistic and unconstrained outdoor environments.

\subsection{More Statistics for LRGait}

As illustrated in Fig.~\ref{supfig::weathers_variables}, we provide detailed statistics of the LRGait dataset. Specifically, we report the number of subjects recorded under various weather conditions, as well as the distribution of activity-related variations such as backpacks, suitcases, and handheld objects across all captured frames.

\begin{figure*}[t]
\centering
\includegraphics[width=\textwidth]{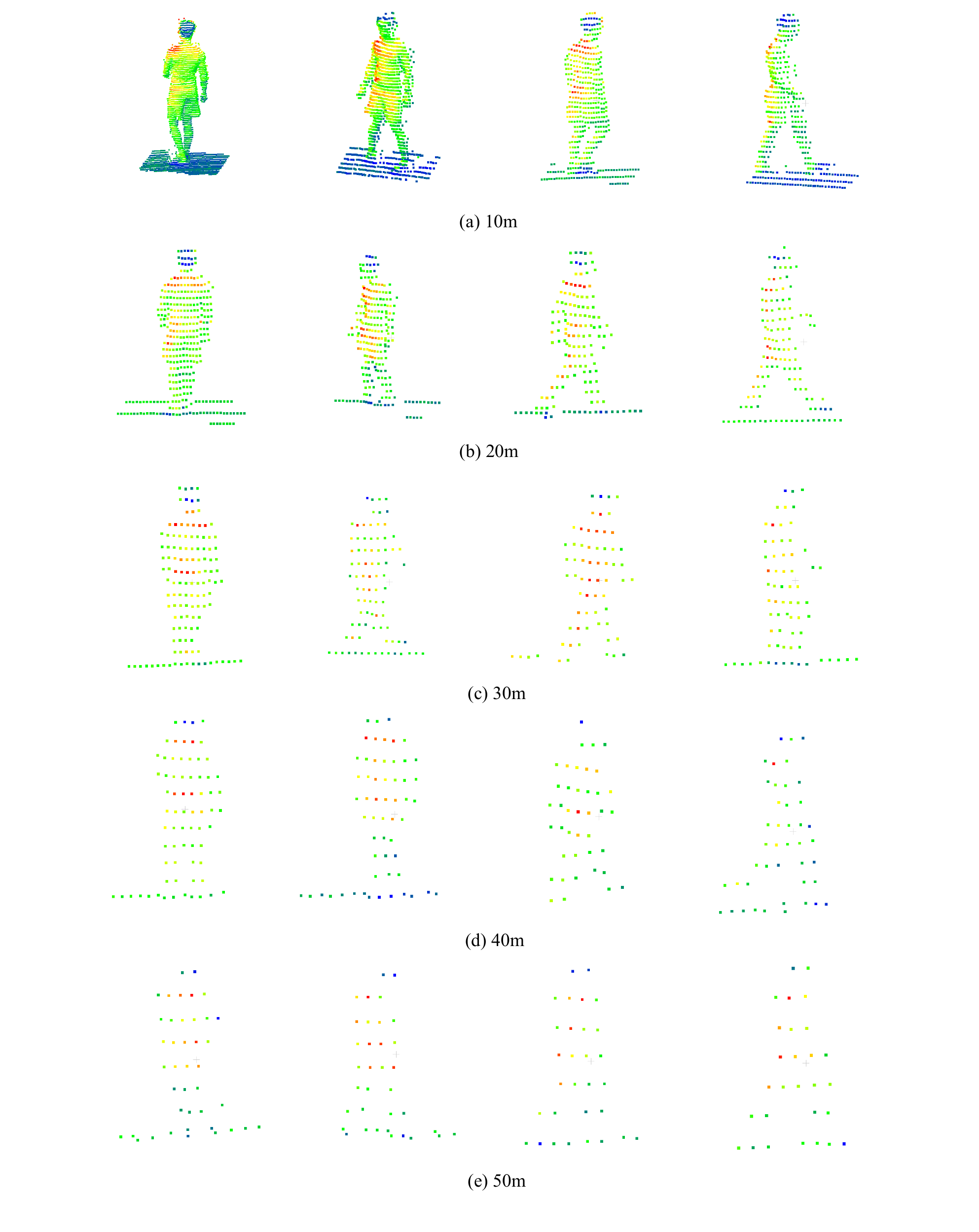}
\caption{10-50m pedestrian point cloud data.}
\label{supfig::pc2}
\end{figure*}

\begin{figure*}[t]
\centering
\includegraphics[width=\textwidth]{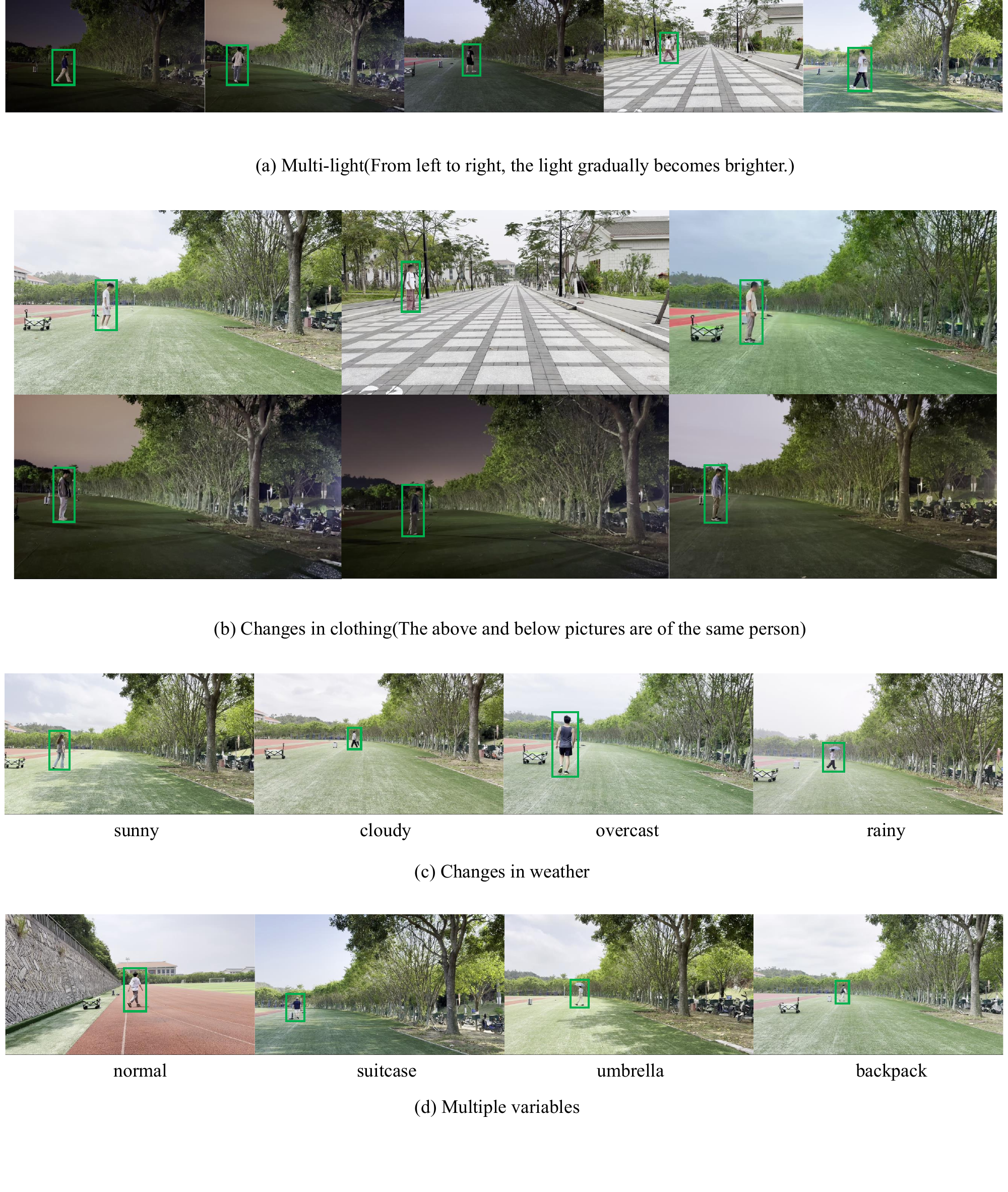}
\caption{Diversity and challenges of LRGait data.}
\label{supfig::details}
\end{figure*}

Overall, LRGait encompasses three distinct scenes, four types of weather conditions, five discrete distance levels (ranging from 10$m$ to 50$m$), and a wide array of real-world behavioral variations. Fig.~\ref{supfig::pc2} highlights the increasing sparsity of LiDAR point clouds with longer distances, illustrating the growing difficulty of reliable gait recognition under extended range. In addition, Fig.~\ref{supfig::details} showcases the diversity, complexity, and real-world fidelity of LRGait, further demonstrating its value as a challenging and comprehensive benchmark for long-range multimodal gait recognition.


\subsection{Dataset Split and Protocols}
All ablation experiments are conducted on the LRGait dataset, which is divided into two non-overlapping splits: a training set comprising 50 identities and 2,800 sequences, and a test set containing the remaining 51 identities and 2,480 sequences. Among them, 20 subjects in the training set and 11 subjects in the test set have gait data captured under both daytime and nighttime conditions.
\bibliography{aaai2026}

\end{document}